\documentclass[runningheads]{llncs}
\usepackage{graphicx}
\usepackage{amsmath}
\usepackage{hyperref}
\usepackage{float}
\usepackage{makecell}
\usepackage{amssymb}
\begin{document}
\title{Sparse 3D Point-cloud Map Upsampling and Noise Removal as a vSLAM Post-processing Step: Experimental Evaluation}
\titlerunning{Sparse 3D Point-cloud Map Upsampling and Noise Removal ...}
\author{Andrey Bokovoy\inst{1,2} \and
Konstantin Yakovlev\inst{2,3}}
\authorrunning{A. Bokovoy and K. Yakovlev.}
\institute{Peoples’ Friendship University of Russia (RUDN University) \\
\email{1042160097@rudn.university}  
\and
Federal Research Center "Computer Science and Control"
of Russian Academy of Sciences \\
\email{\{bokovoy,yakovlev\}@isa.ru}
\and
National Research University Higher School of Economics\\
\email{kyakovlev@hse.ru} }
\maketitle              % typeset the header of the contribution
\begin{abstract}
The monocular vision-based simultaneous localization and mapping (vSLAM) is one of the most challenging problem in mobile robotics and computer vision. In this work we study the post-processing techniques applied to sparse 3D point-cloud maps, obtained by feature-based vSLAM algorithms. Map post-processing is split into 2 major steps: 1) noise and outlier removal and 2) upsampling. We evaluate different combinations of known algorithms for outlier removing and upsampling on datasets of real indoor and outdoor environments and identify the most promising combination. We further use it to convert a point-cloud map, obtained by the real UAV performing indoor flight to 3D voxel grid (octo-map) potentially suitable for path planning. 

\end{abstract}
\section{Introduction} \label{introduction}
%tuna2014autonomous
Simultaneous localization and mapping (SLAM) is a well-known problem in mobile robotics, which is is considered for a variety of different applications~\cite{Cadena16tro-SLAMfuture,balan2015distributed} and platforms~\cite{li2014lidar,leonard2016autonomous}, with unmanned aerial or ground vehicles being the most widespread robots to use SLAM as part of the navigation loop~\cite{caballero2009vision,sazdovski2011inertial,vu2017algorithms}. There exists no universal SLAM method suitable for all robotic platforms and applications due to limitations these platforms/applications impose. Among the factors that influence SLAM the most one can name the following: available data (which in turn depends on the sensors type) and available computing capacities. One of the most challenging scenarios for SLAM is when only video-data, obtained from a single camera, is available and computational resources are limited. This is a typical scenario for UAV navigation, and it leads to so-called monocular vision-based SLAM (vSLAM)~\cite{buyval2017icmv}. vSLAM methods rely on the single-camera video-flow to construct (preferably in real-time) consistent 3D map of the unknown environment and can be classified into 2 major groups: indirect (or feature-based) and direct (dense and semi-dense) methods. 

%bay2006surf,ke2004pca
Indirect vSLAM algorithms utilize images' features~\cite{rublee2011orb} for mapping purposes. Thus the obtained map consists of the set of reconstructed image-features appropriately placed in 3D space. Since the amount of such features for every image is limited and is far less than image's size, the reconstructed map is likely to be sparse and contain large amount of free space (which is actually not free w.r.t obstacles) between the features. On the other hand, most of the feature-detectors utilized in vSLAM work fast (achieving real-time performance) and are invariant to image distortions, light, scale, rotation, etc., which makes them well-suited for real-world robotics applications%~\footnote[1]{\url{https://www.youtube.com/watch?v=piuVq8f61gs}}
.
%\begin{figure}[H]
%\includegraphics[width=\textwidth]{img/fig1.eps}
%\caption{The demonstration of the reconstructed map of the environment (a) with indirect (b) and direct methods} %\label{fig1}
%\end{figure}

Direct methods, like LSD-SLAM~\cite{engel2014lsd,bokovoy2018icarob} or D-TAM~\cite{newcombe2011dtam} use the entire images to reconstruct the map, leading to dense (or semi-dense) maps with large amount of environment details captured. These methods are sensitive to the input data, e.g. they can not handle well distortions, rolling shutter and other typical noise disturbances. They can not run in real-time (without GPU acceleration) as well. One should also mention direct vSLAM methods based on machine learning techniques (e.g. convolutional neural networks), see \cite{tateno2017cnn,makarov2017mm} for example, that have appeared recently. Unfortunately, they require significant computational resources and time to learn before application.

Obviously an ideal vSLAM method should combine the strengths of both approaches, e.g. it should construct detailed maps like direct algorithms do and be fast and robust like indirect ones are. In order to achieve such performance, we suggest to post-process sparse maps, produced by feature-based vSLAM algorithms in order to make them more detailed and suitable for solving further navigation tasks (like path planning~\cite{yakovlev2015ki,magid2011building,makarov2016smoothing}, control~\cite{buyval2017iros} etc.). Such an approach potentially leads to producing detailed maps of the environment with no extra computing costs associated with running direct vSLAM methods.

In this work we study different post-processing techniques, e.g. outlier removal and upsampling, applied to 3D sparse point-cloud maps generated by state-of-the-art feature-based vSLAM algorithms. We evaluate different techniques on various datasets (indoor and outdoor) to find the best combination. We further use it to construct the octo-map of the indoor environment which was not the part of the training datasets.

%The remaining of the paper is organized as follows. Formal definition of vSLAM problem and map post-processing problem are given in  Sect.~\ref{problemstatement}. In Sect.~\ref{methodsandalgorithms} evaluated methods and algorithms are described. Sect.~\ref{experiment} describes the techniques used for the evaluation as well as presents results of experimental study. Sect.~\ref{conclusion} concludes.

\section{Problem Statement} \label{problemstatement}

\subsection{vSLAM problem definition}

The vision-based simultaneous localization and mapping problem for monocular camera (monocular vSLAM) is defined as follows. Let the matrix $I_t \in \mathbb{R}^{m\times n}$ denote the image of $m \times n$ pixels, obtained by the robot at time step $t$\footnote[1]{For the sake of simplicity we assume that the image is grayscale and pixels are real numbers.}. Thus the video-flow $\mathbf{I}_T$ is the sequence $\mathbf{I}_T = \{I_t \; |\; t \in [1,T]\}$,
%T \in \mathbb{N}$,
where $T$ is the end-time.

Given $\mathbf{I}_T$, the localization task is to compute positions of the camera in the global coordinate frame: $\mathbf{X}_T=\{\mathbf{x}_t \; |\; \mathbf{x}_t = (x, y, z, \alpha, \beta, \gamma)\}$,
%x, y, z \in \mathbb{R}, \alpha, \beta, \gamma \in \mathbb{R}$, 
where $x,y,z$ are translation coordinates and $\alpha, \beta, \gamma$ are the orientation angles (e.g. pitch, roll and yaw). 

Furthermore, for each $I_t$ we need to find a set of image points $P_t=\{p_i \; |\; i \leqslant K \leqslant m \times n\}, P_t \in 2^{I_t}$, such that $P_t = f_{proj}( E, t )$, where $E = \{ e_l \; |\; e_l \in \mathbb{R}^3\}$ is the environment and $f_{proj}$ is the function, that projects 3D points from current observation $( E, t )$ to the 2D image $I_t$ as $P_t$.

Finally, the map $\mathbf{M}$ should be constructed using all the observations:

\begin{gather} \label{eq:4}
    M_t = \{f_{proj}^{-1}(p_i) \; |\; i \leqslant m \times n, p_i \in P_t\} \notag \\
    \mathbf{M} = \bigcup\limits_{t=1}^{T}M_t, \\
    \mathbf{M} = \{m_i \; |\; m_i \in \mathbb{R}^3\}. \notag
\end{gather}

\subsection{Map post-processing problem definition}

Consider a filter that is function $filt:\mathbb{R}^3 \to 2^{\mathbb{R}^3}$ and a post-processed map, $\mathbf{\widehat{M}}$, that is constructed by sequentially applying the limited number of filters to the initial map: $\mathbf{\widehat{M}} = filt_1 \circ filt_2 \circ \dotso \circ filt_H (\mathbf{M})$.

We now want to find such combination of filters that enriches the map with additional points (the map becomes more dense) and at the same time keeps the model as close to the ground-truth as possible. Formally, $|\mathbf{\widehat{M}}|$ should be maximized and $Error( \mathbf{\widehat{M}}, E )$ should be minimized, where $Error( \mathbf{\widehat{M}}, E) = \frac{1}{|\mathbf{\widehat{M}}|} \sum_{r=1}^{|\mathbf{\widehat{M}}|} \left \lVert \widehat{m}_{r} - e_r \right \lVert $, $m_r \in \mathbf{\mathbf{M}}$, $e_r = corr( m_r )$ -- correspondence function between $E$ and $\mathbf{\widehat{M}}$ (Euclidean distance, for example).

%Post-processing is needed to enhance the quality of the map, constructed by a vSLAM algorithm, that is: given $\mathbf{M}$ a set of filters, $filt_j:\mathbb{R}^3 \to 2^{\mathbb{R}^3}$, should be found:

%\begin{equation}
%    \mathbf{\widehat{M}} = filt_1 \circ filt_2 \circ \dotso \circ filt_H (\mathbf{M}) 
%\end{equation}

%Lets denote the post-processed map as $\mathbf{\widehat{M}}$. Considering the map $M$ from equation 1, we need to find a combination of functions $filt_j:\mathbb{R}^3 \to \mathbb{R}^3, j \in \mathbb{N}$:

%that maximizes $|\mathbf{\widehat{M}}|$ and minimizes the error between $\mathbf{\widehat{M}}$ and $\mathbf{M}$ -- $e = Error( \mathbf{\widehat{M}}, \mathbf{M} ) = \frac{1}{|\mathbf{\widehat{M}}|} \sum_{r=1}^{|\mathbf{\widehat{M}}|} \left \lVert \widehat{m}_{r} - m_{r} \right \lVert $, $\widehat{m}_r \in \mathbf{\widehat{M}}$, $m_r \in \mathbf{\mathbf{M}}$. Here $|\mathbf{\widehat{M}}|$ denotes the enhanced point-cloud map of the environment.

\section{Evaluated Methods and Algorithms} \label{methodsandalgorithms}

We need to choose a suitable vSLAM method, which is able to produce maps that can be further used (possibly after the described post-processing phase) for various navigation tasks, with path planning being of the main interest. This method should be applicable to real-world robotic applications, e.g. it should be i) fast (able to process at least $640 \times 480$ grayscale images at 30 Hz), ii) able to work with distorted images, iii) well studied and it's implementation should be available for the community. Based on the these criteria and taking into account the considerations specified in section \ref{introduction}, ORB-SLAM2~\cite{mur2017orb2} was chosen. We also took into account the evaluation results of~\cite{webKITTI}.

%provided by The KITTI Vision Benchmark Suite~\cite{Geiger2012CVPR}. 

\begin{figure}[H]
\includegraphics[height=5cm]{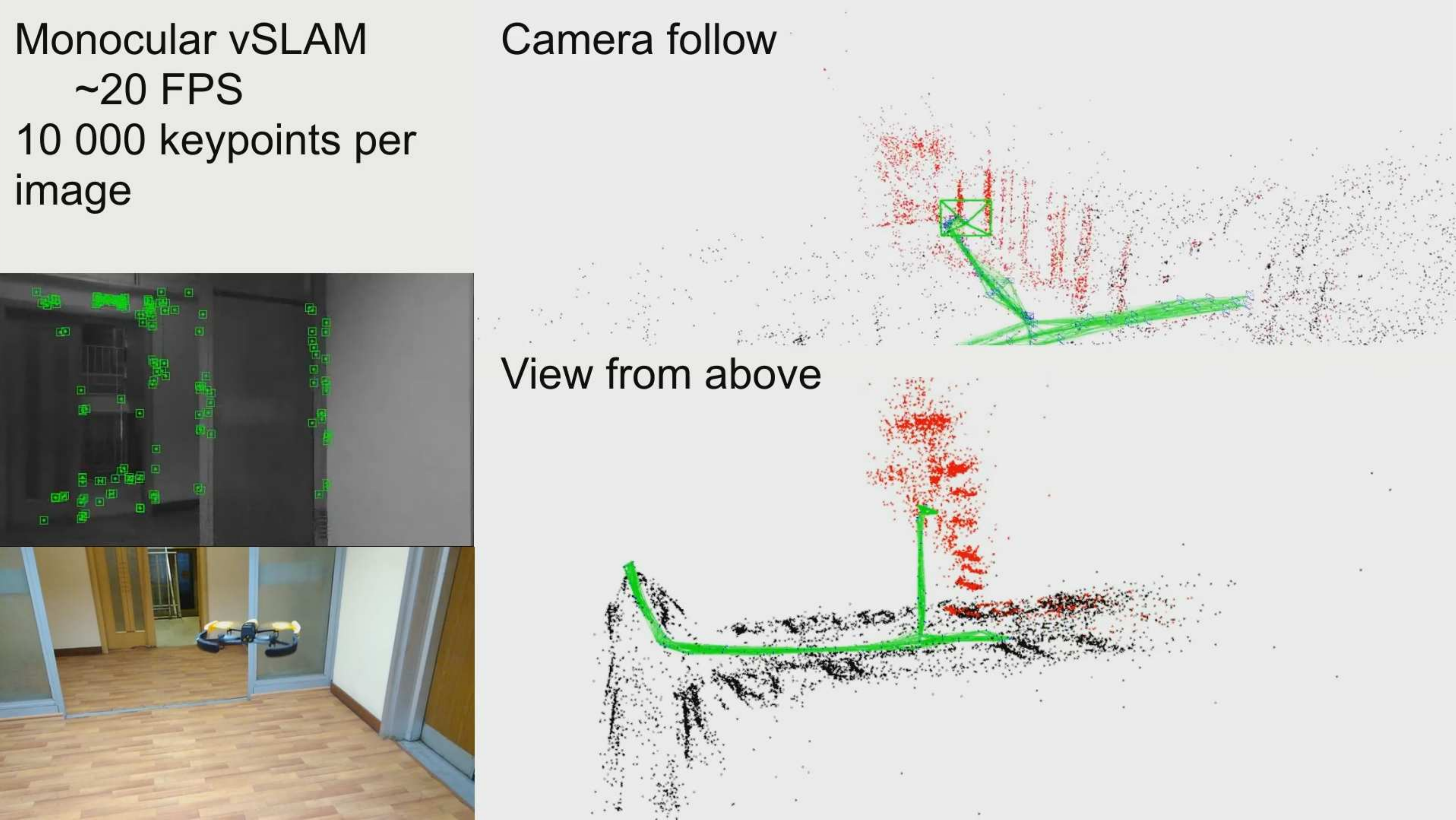}
\centering
\caption{ORB-SLAM2 output on real-world indoor flight performed by Parrot Bebop quadrotor. Video is available at \url{https://www.youtube.com/watch?v=piuVq8f61gs} } \label{fig2}
\end{figure}

The results of running ORB-SLAM2 on real data collected by the compact quadcopter, performing its flight in the indoor environment of our institute, are shown on \figurename ~\ref{fig2}.

%and the reconstructed 3D grid map~\cite{hornung2013octomap}

As one can see on \figurename \ref{fig2}, ORB-SLAM2 map is sparse and noisy. To make the output more suitable for further conversion to the octomap (3D grid) \cite{hornung2013octomap} we suggest applying 2 following steps: i) outlier removal, ii) upsampling.

%Thus (considering equation 5) we need to find $filt_1$ and $filt_2$, where $filt_1$ is outlier removal filter and $filt_2$ - upsampling filter. Further, we examine both steps in details.

\subsection{Outlier removal}

There exists 2 general approaches to outlier removal for point-clouds: radius-based and statistical~\cite{rusu2008towards}. Radius-based methods filters the elements of the point-cloud based on the amount of neighbors they have. It iterates over input point-cloud once and retrieves the number of neighbors within the certain radius $r$. If this number is less than the predefined threshold $b$ the point is considered an outlier.

Statistical approaches iterate throw the input point-cloud twice. During the first iteration average distance from each point to its nearest $l$ neighbors is estimated. Consequently, the mean and standard deviation are computed for all the distances in order to determine a threshold. On the second iteration the points will be considered as outliers if their average neighbor distance is above this threshold. The main parameter of statistical methods is the standard deviation multiplier $h$ that affects the final threshold. 

\subsection{Upsampling}
   
Almost all upsampling filters for point-clouds are based on Moving Least Squares (MLS)~\cite{reuter2005surface} techniques. This techniques involve the projecting of the point-clouds into continues surface that minimizes a local least-square error. We choose the most common upsampling methods, such as Sample Local Plane, Random Uniform Density and Voxel Grid Dilation~\cite{skinner20143d} for further evaluation. These methods are parameter-dependent and the parameters are: i) upsampling radius ($u_r$), upsampling step size ($u_{sz}$) and maximum number of upsampling steps ($u_s$) for Sample Local Plane, ii) point density ($d$) for Random Uniform Density, iii) dilation voxel size ($s_{vs}$) and dilation iterations ($d_i$) for Voxel Grid Dilation.
    
\subsection{Map scaling}

For upsampling and oulier removal methods we need to find the best parameters at which this algorithms produce the most accurate and detailed maps. For this purposes, we need to compare the output of each method and their combinations with ground truth. Since ORB-SLAM2 produces the maps with unknown scaling, we need find corresponding points (algorithm 1) and adjust the scaling of ORB-SLAM2's map and ground truth (algorithm 2). 

\begin{table}[H]
\centering
\begin{tabular}{p{10cm}}
\hline
  \textbf{Algorithm 1} Corresponding points search. \\
\hline

\begin{enumerate}
\item Get $\mathbf{M}, \mathbf{X}_T, P_T = \{P_i\}, i \in [1,T]$ for ground-truth and $\mathbf{\widehat{M'}}, \mathbf{X'}_T$ for post-processed map.
\item Get $\mathbf{I}_T$ with corresponding ground-truth $E_T$ 
\item \textbf{for} each $I_i \in \mathbf{I}_T$
\item \quad \textbf{for} each pix $pix \in I_i$
\item \quad \quad \textbf{if} $pix \in f_{proj}(E_i)$
\item \quad \quad \quad find the correspondence $p'$ for $pix$ in $\mathbf{\widehat{M'}}$ if exists
\item \quad \quad \quad add $p'$ to $P'_i$
\item \quad \quad \textbf{end if}
\item \quad \textbf{end for}
\item \quad \textbf{for} each $m_i \in \mathbf{M}$ with $m'_i \in \mathbf{\widehat{M'}}$ with correspondences $P'_i$ 
\item \quad \quad Calculate per coordinate deviation: $D_i = m_i - \widehat{m'}_i$
\item \quad \quad Add $D_i$ to $D$
\item \quad \textbf{end for}
\item \textbf{end for}
\item \textbf{return} $D$
\end{enumerate} \\
\hline
\end{tabular}
\end{table}

\begin{table}[H]
\centering
\begin{tabular}{p{10cm}}
\hline
  \textbf{Algorithm 2} Map scaling. \\
\hline

\begin{enumerate}
\item Get an the resultant map $\mathbf{\widehat{M'}}$ with corresponding trajectory $\mathbf{X'}_T$ 
\item For ground truth map $\mathbf{M}$ and trajectory $\mathbf{X}_T$ adjust the $x_1$ pose to $x'_1$, $x_T$ to $x'_T$ and $x_l$ with $x'_l$, where $l \in (1,T), l \in N$
\item Find the scale factor $s=(s_x,s_y,s_z)$ for each coordinate $x,y,z$ using translation of the poses $\mathbf{X'}_T$
\item \textbf{return} $\mathbf{\widehat{M}}=\{s\circ m'_{pp_i}, i \in [1,K]\}$
\end{enumerate} \\
\hline
\end{tabular}
\end{table}
    
\section{Experimental Analysis and Results} \label{experiment}

Experimental evaluation consists of 2 main stages: \textbf{parameters adjustment} and \textbf{map quality estimation}. During the first stage we used limited amount of input data to adjust the parameters of outlier removal and upsampling filters. We also searched for best combination of the upsampler and the outlier removal. On the second step, we extrapolated the estimated parameters to a large variety of input data to estimate the quality of post-processed map. After all, we evaluated the suggested pipeline on the real-world scenario depicted in \figurename \ref{fig2}. 

\subsection{Tools}

We used open-source realization of ORB-SLAM2~\footnote[2]{\url{https://github.com/raulmur/ORB_SLAM2}}, provided by its authors for sparse map construction and PointCloud Library (PCL)~\cite{rusu20113d} with built-in implementations of upsampling and outlier removal algorithms for map enhancement. Experiments were run on the 3-PC cluster with each experiment executed in it's own processor's thread. 

\begin{figure}[H]
\includegraphics[width=\textwidth]{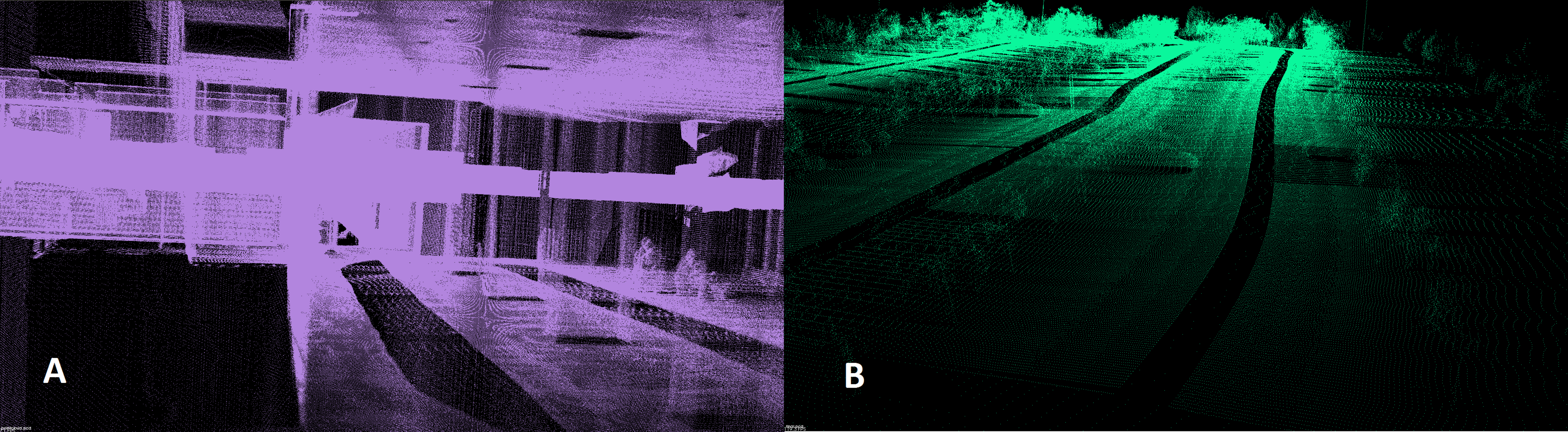}
\centering
\caption{Datasets, used for experimental evaluation. (a) TUMindoor Dataset (b) Malaga Dataset 2009} \label{fig3}
\end{figure}

2 datasets were used: TUMindoor Dataset~\cite{huitl2012tumindoor} and Malaga Dataset 2009~\cite{blanco2009collection}. Malaga Dataset 2009 consists of 6 outdoor environments with ground-truth map, 6-DOF camera poses and corresponding video sequences. TUMidoor Dataset consists of the sequences, gathered inside of Technische Universität München with ground-truth map and 6-DOF camera poses. The path length varies from 120m to 1169m. We split each initial sequence form the datasets into smaller sequences with a fixed path length of 10m, thus 45 sequences from Malaga Dataset and 65 sequences from TUMindoor Dataset were used.
The example of the provided environment is shown in \figurename \ref{fig3}.

\subsection{Parameters adjustment}
To adjust the parameters we used TUM RGBD-SLAM Dataset and Benchmark~\cite{sturm2012benchmark}, particularity the ``freiburg2\_desk\_validation" sequence.

We varied each parameter described in section~\ref{methodsandalgorithms} for each of the upsampling and outlier removal method. 25 000 of the parameters' combinations were evaluated in total. We estimated the runtime, resultant map size and map's accuracy compared to ground-truth. The results for best parameterizations are shown in Table~\ref{tab1}. As one can see, the best performance is achieved by statistical outlier with $h=1.8$ and by Voxel Grid Dilation with dilation voxel size set to 4.9 and dilation iterations set to 3.

\begin{table}
\caption{Map accuracy and processing time.}\label{tab1}
\centering
\begin{tabular}{|l|l|l|l|l|}
\hline
Algorithm &  Parameters & \makecell{Deviation in \% \\ (compared to \\ ground-truth)} & Time (s) & \makecell{Map \\ points}\\
\hline
ORB-SLAM2 & - & 2.81 & - & 11 546 \\
\hline
Radius filter &  \makecell[l]{$b=4$ \\ $r=8.9$} & 2.34 & 4 & 85 982\\
\hline
Statistical filter &  $h=1.8$ & 2.01 & 1.52 & 87 449\\
\hline
Sample Local Plane & \makecell[l]{$u_r=1.12$ \\ $u_sz = 0.58$ \\ $u_s=118$ } & 1.98  & 3.2 & 90 178\\
\hline
Random Uniform Density &  $d=13$ & 2.13 & 4 & 89 965\\
\hline
Voxel Grid Dilation &  \makecell[l]{$s_{vs}=4.9$ \\ $d_i=3$}  & 1.83 & 2.1 & 95 676\\
\hline
\end{tabular}
\end{table}

\subsection{Map Quality Estimation}

\begin{figure}
\includegraphics[width=8cm]{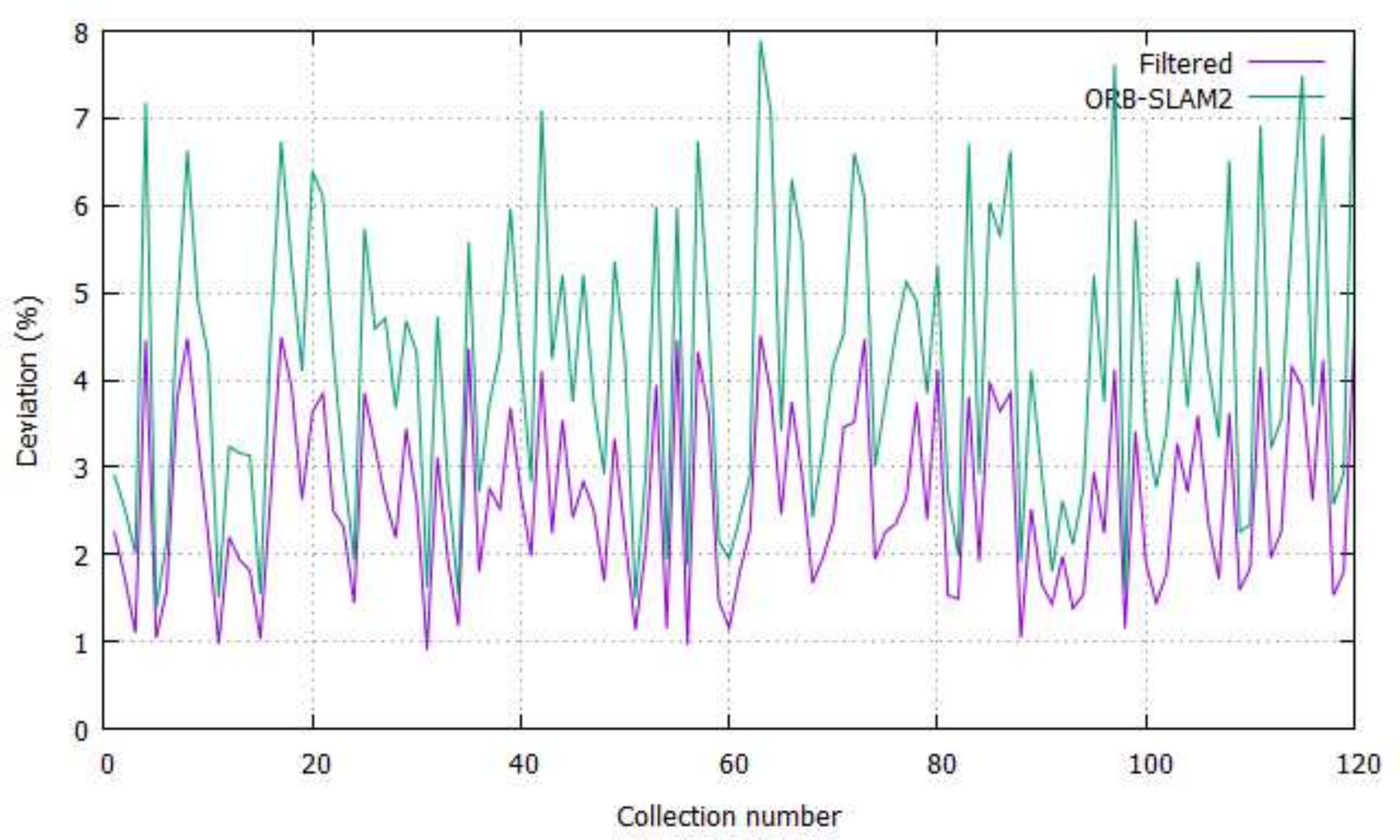}
\centering
\caption{Average deviation of the post-processed map and ORB-SLAM2's map from ground-truth. Less is better.} \label{fig5}
\end{figure}

We combined the statistical outlier filter with Voxel Grid Dilation upsampling algorithm to post-process the maps obtained by ORB-SLAM2 on all the available data: 110 data instances from both TUMidoor Dataset and Malaga Dataset. The results of the evaluation are shown in \figurename \ref{fig5}. As one can see, the suggested approach is able to produce more precise and dense maps, compared to original ORB-SLAM2.

\begin{figure}
\includegraphics[width=\textwidth]{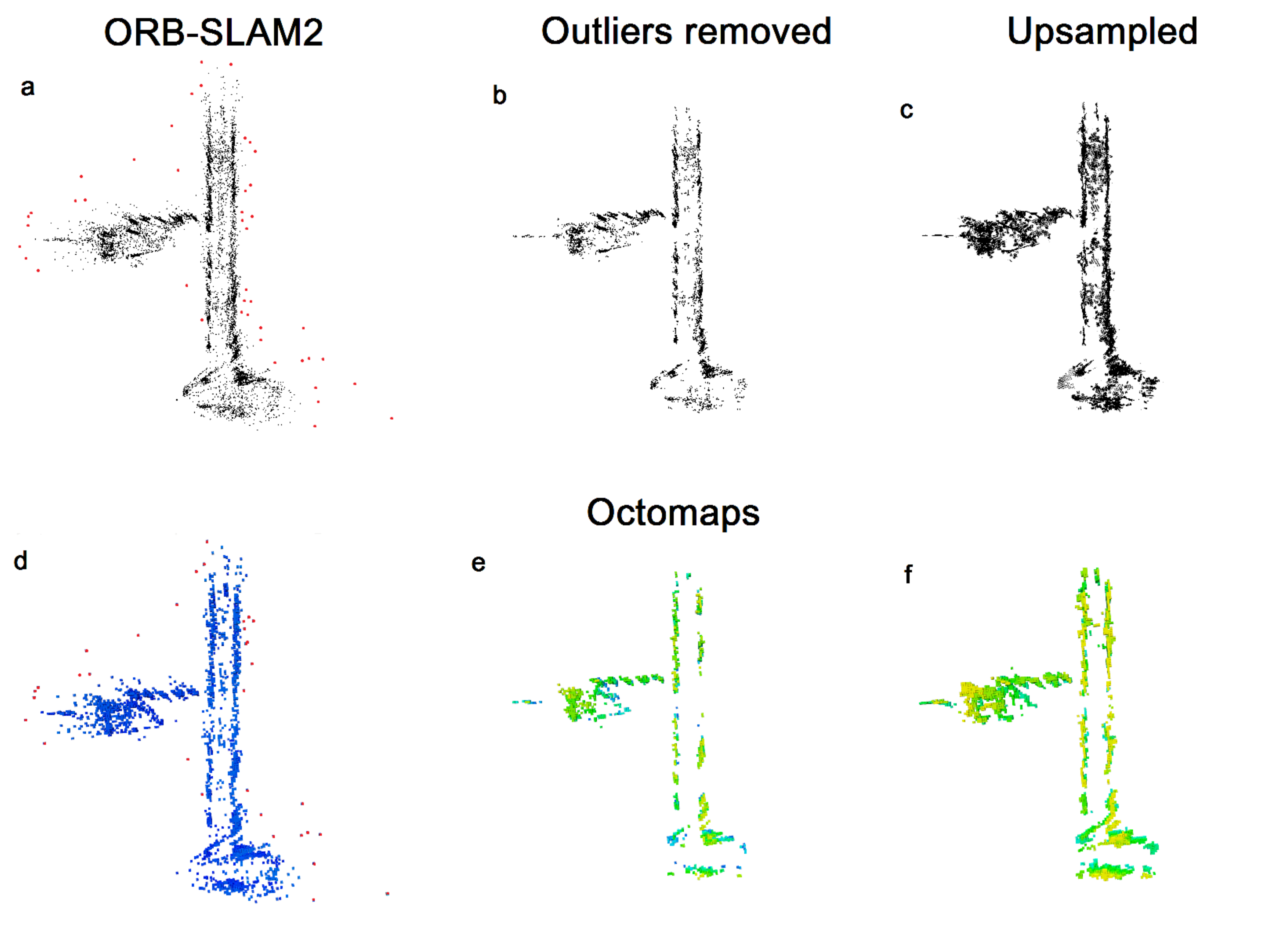}
\centering
\caption{a) Original map, produced by ORB-SLAM2 (outliers are highlighted in red); b) the map with outliers removed; c) upsampled map. d), e), f) --  corresponding octrees (with the ceiling and the floor removed for better visualization).} \label{fig6}
\end{figure}

Finally we tested the suggested pipeline on the video-data, captured form Bebop quadrotor performing indoor flight in our lab (video is available at \url{https://www.youtube.com/watch?v=piuVq8f61gs}) The visualization of the result is given on \figurename~\ref{fig6}. Original map contains 8838 points, smoothed map has 6832 and upsampled map - 58416. As one can see, applying the suggested outlier removal and upsampling filters positively influence the quality of the resultant point-cloud map and, as a result, converted octo-map becomes more suitable for further usage (e.g. for path planning).

\section{Conclusion} \label{conclusion}

We have considered the problem of enhancing the maps produced by monocular feature-based vSLAM (ORB-SLAM2). This problem naturally arises in various mobile robotics applications as typically the feature-based vSLAM maps are extremely sparse. We evaluated the post-processing pipeline that includes outlier removal and upsampling. Different combinations of known methods were evaluated and the best parameters for each method were identified. The best combination was then extensively tested on both well-known in the community indoor and outdoor collections of video-data and the video from real quadrotor captured in our lab. The results of such evaluation showed the increase of the accuracy and the density of the post-processed maps.

\bigskip
\subsubsection*{Acknowledgment}
This work was partially supported by the ``RUDN University Program 5-100'' and by the RFBR project No. 17-29-07053.

\bibliographystyle{splncs}
\bibliography{bibl.bib}

\end{document}